\documentclass[11pt,a4paper]{article}
\usepackage[hyperref]{acl2021}
\usepackage{times}
\usepackage{latexsym}
\usepackage{url}
\usepackage{color}
\usepackage{multirow}
\usepackage{amsmath}
\usepackage{graphicx}
\usepackage{subfig}
\usepackage{rotating}

\usepackage{microtype}

\aclfinalcopy 
\def\aclpaperid{9} 

\title{A Template-guided Hybrid Pointer Network for Knowledge-based Task-oriented Dialogue Systems}

\author{Dingmin Wang$^{1}$, ~ Ziyao Chen$^{2}$, ~Wanwei He$^{3}$, ~Li Zhong$^{4}$, Yunzhe Tao$^{5}$, Min Yang$^{3}$\\
$^{1}$Department of Computer Science, University of Oxford, UK\\
$^{2}$Tencent, Guangzhou, China\\
$^{3}$Shenzhen Institutes of Advanced Technology, Chinese Academy of Sciences, China\\
$^{4}$ByteDance, Shenzhen, China,   $^{5}$Amazon Web Services, Seattle, USA \\
\texttt{dingmin.wang@cs.ox.ac.uk}~~~~\texttt{yateschen@tencent.com}~~~~\texttt{yunzhet@amazon.com}\\ \texttt{\{ww.he,min.yang\}@siat.ac.cn}~~~~\texttt{zhongli.reginald@bytedance.com}\\
}
\begin{document}
\maketitle
\begin{abstract}
Most existing neural network based task-oriented dialogue systems follow \textit{encoder-decoder} paradigm, where the decoder purely depends on the source texts to generate a sequence of words, usually suffering from instability and poor readability. Inspired by the traditional template-based generation approaches, we propose a template-guided hybrid pointer network for the knowledge-based task-oriented dialogue system, which retrieves several potentially relevant answers from a pre-constructed domain-specific conversational repository as guidance answers, and incorporates the guidance answers into both the encoding and decoding processes. Specifically, we design a memory pointer network model with a gating mechanism to fully exploit the semantic correlation between the retrieved answers and the ground-truth response. We evaluate our model on four widely used task-oriented datasets, including one simulated and three manually created datasets. The experimental results demonstrate that the proposed model achieves significantly better performance than the state-of-the-art methods over different automatic evaluation metrics~\footnote{\url{https://github.com/wdimmy/THPN}}.
\end{abstract}

\begin{table*}[t]
\def\arraystretch{1.3}
\centering\small
  \caption{Two example conversations from real dialogues about navigation and weather. \label{tab:extracts}}
  \begin{tabular}{p{1.0cm} | p{6.1cm} | p{1.0cm} | p{6.1cm}} 
    \hline
    \multicolumn{2}{c|}{\textbf{Navigation} }  & \multicolumn{2}{c}{\textbf{Weather} } \\
    \hline 
    \textbf{User} & please give me directions to \textcolor{blue}{5677\_spring\_street} &\textbf{User} & what is the temperature of \textcolor{blue}{carson} on tuesday \\
    \hline
    \multirow{2}{*}{ Retrieve} & q1: direct me to stanford children's health &  \multirow{2}{*}{ Retrieve } & q1: the temperature of new\_york on wednesday\\
        & a1:  no problem, I will be navigating you to stanford children's health right now &  
        & a1:   the temperature in new\_york on wednesday  will be  low\_of\_80f  and high\_of\_90f\\
    \hline 
   \multirow{2}{*}{ KB}  &   &   \multirow{2}{*}{ KB}  &  carson: tuesday \textcolor{red}{low\_of\_20f}   \\
   &  &  & carson: tuesday \textcolor{red}{high\_of\_40f} \\
   
    \hline 
        \textbf{Gold} &  no problem, I will be navigating you to \textcolor{blue}{5677\_spring\_street} right now & \textbf{Gold} &  the temperature in \textcolor{blue}{carson} on tuesday will be \textcolor{red}{low\_of\_20f} and \textcolor{red}{high\_of\_40f}\\
                             
    \hline 
      
   \hline 
  \end{tabular}
\end{table*}

\section{Introduction}
Task oriented dialogue systems have attracted increasing attention recently due to broad applications such as reserving restaurants and booking flights. Conventional task-oriented dialogue systems are mainly implemented by  rule-based methods~\cite{lemon2006isu,wang2013simple}, which rely heavily on the hand-crafted features, establishing significant barriers for adapting the dialogue systems to new domains. Motivated by the great success of deep learning in various NLP tasks, the neural network based methods~\cite{bordes2017learning,eric2017copy,madotto2018mem2seq} have dominated the study since these methods can be trained in an end-to-end manner and scaled to different domains.

Despite the remarkable progress of previous studies, the performance of task-oriented dialogue systems is still far from satisfactory. On one hand, due to the exposure bias problem \cite{ranzato2015sequence}, the neural network based models, e.g., the sequence to sequence models (seq2seq), tend to accumulate errors with increasing length of the generation. Concretely, the first several generated words can be reasonable, while the quality of the generated sequence deteriorates quickly once the decoder produces a ``bad'' word. On the other hand, as shown in previous works~\cite{cao2018retrieve,madotto2018mem2seq}, the Seq2Seq models are likely to generate non-committal or similar responses that often involve high-frequency words or phrases. These responses are usually of low informativeness or readability. This may be because that arbitrary-length sequences can be generated, and it is not enough for the decoder to be purely based on the source input sentence to generate informative and fluent responses.

We demonstrate empirically that in task-oriented dialogue systems, the responses for the requests with similar types often follow the same sentence structure except that different named entities are used according to the specific dialogue context. 
Table~\ref{tab:extracts} shows two conversations from real task-oriented dialogues about navigation and weather. From the navigation case, we can observe that
although the two requests are for different destinations, the corresponding responses are similar in sentence structure, replacing ``children's health'' with ``5677\_springer\_street''.
For the weather example, it requires the model to first detect the entity ``carson'' and then query the corresponding information from the knowledge base (KB). 
After obtaining the returned KB entries, we generate the response by replacing the corresponding entities in the retrieved candidate answer.
Therefore, we argue that the golden responses of the requests with similar types can provide a reference point to guide the response generation process and enable to generate high-quality responses for the given requests.

In this paper, we propose a template-guided hybrid pointer network (THPN to generate the response given a user-issued query, in which the domain specific knowledge base (KB) and potentially relevant answers are leveraged as extra input to enrich the input representations of the decoder. 
Here, \textit{knowledge base} refers to the database to store the relevant and necessary information for supporting the model in accomplishing the given tasks. We follow previous works and use a triple (subject, relation, object) representation. For example, the triple (Starbucks, address, 792 Bedoin St) is an example in KB representing the information related to the Starbucks.
Specifically, given a query, we first retrieve top-$n$ answer candidates from a pre-constructed conversational repository with question-answer pairs using BERT~\cite{devlin2018bert}.
Then, we extend memory networks~\cite{sukhbaatar2015end} to incorporate the commonsense knowledge from KB to learn the knowledge-enhanced representations of the dialogue history. Finally, we introduce a gating mechanism to effectively utilize candidate answers and improve the decoding process.  The main contributions of this paper can be summarized as follows: 
\begin{itemize}
\item We propose a hybrid pointer network consisting of entity pointer network (EPN) and pattern pointer network (PPN) to generate informative and relevant responses. EPN copies entity words from dialogue history, and PPN extracts pattern words from retrieved answers. 
\item We introduce a gating mechanism to learn the semantic correlations between the user-issued query and the retrieved candidate answers, which reduces the ``noise'' brought by the retrieved answers. 
\item We evaluate the effectiveness of our model on four benchmark task-oriented dialogue datasets from different domains. Experimental results demonstrate the superiority of our proposed model.
\end{itemize}

\section{Related Work}

Task-oriented dialogue systems are mainly studied via two different approaches: pipeline based and end-to-end. Pipeline based models \cite{williams2007partially,young2013pomdp} achieve good stability but need domain-specific knowledge and handcrafted labels. End-to-end methods have shown promising results recently and attracted more attention since they are easily adapted to a new domain. 

Neural network based dialogue systems can avoid the laborious feature engineering since the neural networks have great ability to learn the latent representations of the input text. However, as revealed by previous studies~\cite{koehn2017six,cao2018retrieve,he2019quantifying}, the performance of the sequence to sequence model deteriorates quickly with the increase of the length of generation. Therefore, how to improve the stability and readability of the neural network models has attracted increasing attention. \citet{eric2017key}  proposed a copy augmented Seq2Seq model by copying relevant information directly from the KB information. \citet{madotto2018mem2seq} proposed a generative model by employing the multi-hop attention over memories with the idea of pointer network. \citet{DBLP:conf/iclr/WuSX19} proposes a global-to-locally pointer mechanism to effectively utilize the knowledge base information, which improves the quality of the generated response.

Previous proposed neural approaches have shown the importance of  external knowledge in the sequence generation~\cite{chen2017neural,zhu2018retrieval,yang2019enhancing,zhang2019syntax,ding2019event}, especially in the task-oriented dialogue systems where an appropriate response usually requires correctly extracting knowledge from the domain-specific or commonsense knowledge base~\cite{madotto2018mem2seq,zhu2018retrieval,qin2019entity}. However, it is still under great exploration with regard with the inclusion of external knowledge into the model. \citet{yan2016learning,song2018ensemble} argue that retrieval and generative methods have their own demerits and merits, and they have achieved good performance in the chit-chat response generation by incorporating the retrieved results in the Seq2Seq based models. \citet{zhu2018retrieval} proposed an adversarial training approach, which is enhanced by retrieving some related candidate answers in the neural response generation, and \citet{ghazvininejad2018knowledge} also applies a similar method in the neural conversation model. In addition, in task-oriented dialogue tasks, the copy mechanism~\cite{gulcehre2016pointing} 
has also been widely utilized~\cite{eric2017copy,madotto2018mem2seq}, which shows the superiority of generation based methods with copy strategy. 

\section{Methodology}

\begin{figure*}[htb]
  \centering
  \includegraphics[width=0.86\textwidth]{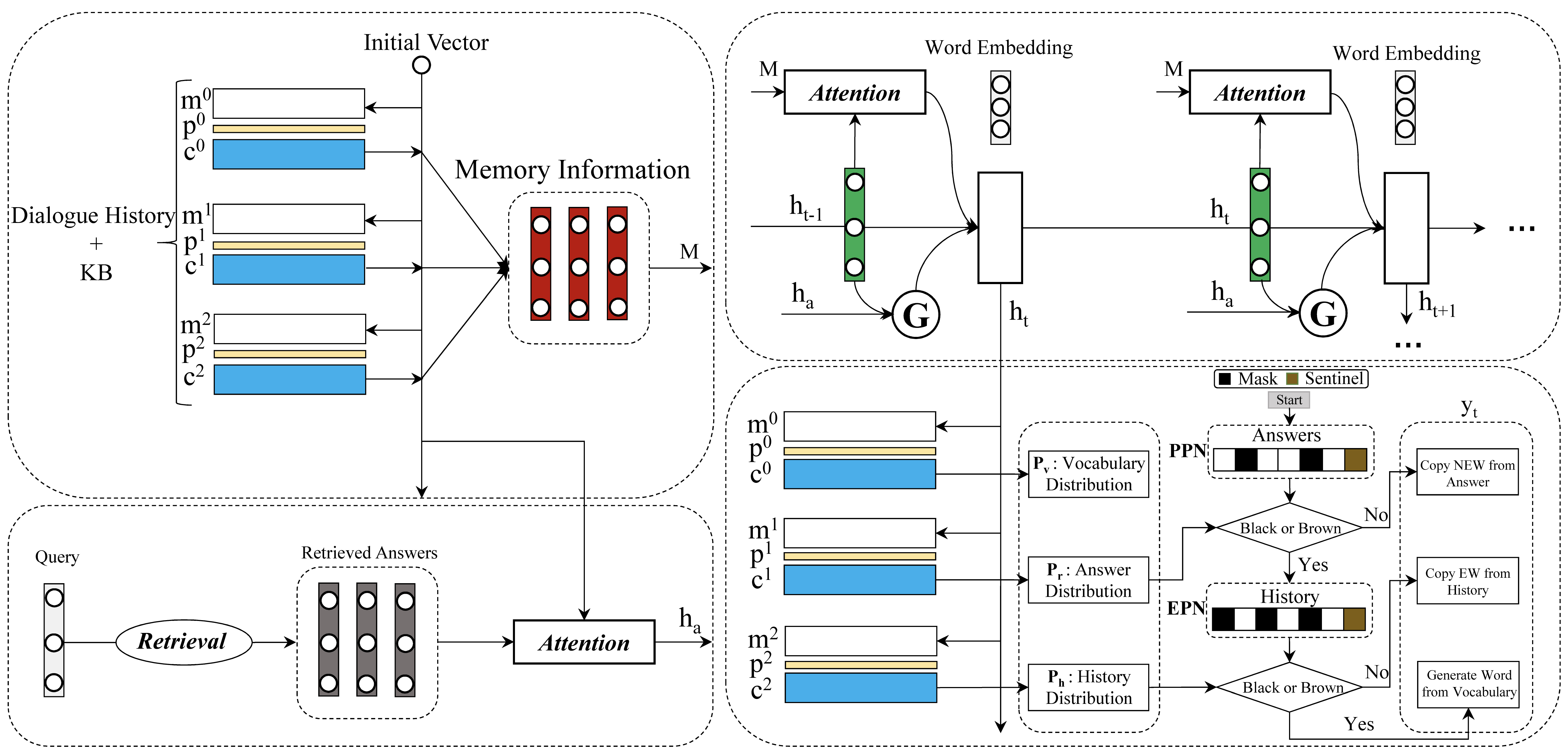}
  \caption{The overall structure of our model.
  During test time, given a user query $q$, we retrieve at most $3$ similar questions to $q$ using BERT from QA Paris repository, and the corresponding answers are used as our answer templates. The retrieved answers as well as the dialogue history and KB information are then utilized for the response generation. Especially, we utilize the gating mechanism to filter out noise from unrelated retrieval results. Finally, words are generated either from the vocabulary or directly copying from the multi-source information using a hybrid pointer network.\label{fig:achitecture}}
\end{figure*}

We build our model based on a seq2seq dialogue generation mode, and the overall architecture is exhibited in Figure~\ref{fig:achitecture}. Each module will be elaborated in the following subsections.

\subsection{Encoder Module}
By checking if a word is in the given KB, we divide words into two types: entity words~(EW) and  non-entity words~(NEW). Taking ``what is the temperature of carson on tuesday'' as an example, all words are NEW except for ``carson'' and ``tuesday''. 

We represent a multi-turn dialogue as $D = \{(u_i, s_i)\}_{i=1}^{T}$, where $T$ is the number of turns in the dialogue, and $u_i$ and $s_i$ denote the utterances of the user and the system at the $i^{th}$ turn, respectively. KB information is represented as $KB=\{k_1, k_2, \cdots, k_l\}$, where $k_i$ is a tuple and $l$ is the size of KB. Following~\citet{madotto2018mem2seq}, we concatenate the previous dialogue and KB as input.
At first turn, input to the decoder is $[u_1;KB]$, the concatenation of first user request and KB. For $i>1$, previous history dialog information is included, namely, input is supposed to be $[u_1, s_1, \cdots, u_i;KB]$. 
We define words in the concatenated input as a sequence of tokens $W = \{w_1, w_2, \cdots, w_n\}$, where $w_j \in \{u_1, s_1, \cdots, u_i, KB\}$ , $n$ is the number of tokens.

In this paper, we use the memory network~(MemNN) proposed in ~\citet{sukhbaatar2015end} as the encoder module. The memories of MemNN are
represented by a set of trainable embedding matrices $M =\{M^1, M^2, \cdots, M^{K}\}$, where $K$ represents the number of hops and each $M^k$ maps the input into vectors. Different from ~\citet{sukhbaatar2015end,madotto2018mem2seq}, we initialize each $M^k$ with the pre-trained embeddings\footnote{\url{https://s3-us-west-1.amazonaws.com/fasttext-vectors/wiki.en.vec}.}, whose weights are set to be trainable. At hop $k$, $W$ is mapped to a set of memory vectors, $\{m_1^k, m_2^k, \cdots, m_{n}^k\}$, where the memory vectors $m_i^k$ of dimension $d$ from $M^k$ is computed by embedding each word in a continuous space, in the simplest case, using an embedding matrix $A$. A query vector $q$ is used as
a reading head, which will loop over $K$ hops and compute the attention weights at hop $k$ for each
memory by taking the inner product followed by a softmax function,
\begin{equation}
p_i^k=\mathit{softmax}\left(\left(q^k\right)^Tm_i^k\right)
\end{equation}
where $p_i^k$ is a soft memory selector that decides the memory relevance with respect to the query vector $q$.
The model then gets the memory $c^k$ by the
weighted sum over $m^{k+1}$,
\begin{equation}
c^k = \sum_i p_i^km_i^{k+1}\
\end{equation}
In addition, the query vector is updated for the next hop
by $q^{k+1} = q^k + c^k$. In total, we can achieve K hidden states encoded from MemNN, represented as $C = \{c^1, c^2, \cdots, c^K\}$.

\paragraph{Masking NEW in the history dialogue} We observe that the ratio of non-entity words in both the history dialogue and the expected response is extremely low. Therefore, to prevent the model from copying non-entity words from the history dialogue, we introduce an array $R_h$\footnote{The length of $R_h$ equals to that of $W$.} whose elements are zeros and ones, where $0$ denotes NEW and $1$ for EW. When $w_i$ is pointed to, and if $i$ is the sentinel location or $R_h[i]=0$, then $w_i$ will not be copied. 

\subsection{Retrieval Module}
For each dataset, we use the corresponding training data to pre-construct a question-answer repository. In particular, we treat each post-response ($u_i$ and $s_i$) in a dialogue as a pair of question-answer. 
To effectively retrieve potentially relevant answers, we adopt a sentence matching based approach,  in which each sentence is represented as a dense vector, and the cosine similarity serves as the selection  metrics. 
We have explored several unsupervised text matching methods, such as BM25~\cite{robertson2009probabilistic}, Word2Vec~\cite{mikolov2013distributed}, and BERT~\cite{devlin2018bert}, and revealed that BERT could achieve the best performance. 
In addition, based on our preliminary experiments, we observed that the number of retrieved answer candidates have an impact on the model performance, so we define a threshold $\theta$ for controlling the number of retrieval answer candidates.

Specifically, for each question in the pre-constructed database, we pre-compute the corresponding sentence embedding using BERT. Then, for each new user-issued query $u_q$, we embed $u_q$ into $u_q^{e}$, and search in the pre-constructed database for the most similar requests based on cosine similarity. The corresponding answers are selected and serve as our answer candidates.

\paragraph{Masking EW in the retrieved answers} In real dialogue scenes, the reply's sentence structure might be similar but the involved entities are usually different. To prevent the model from copying these entities, we introduce another array $R_r$ similar to $R_h$ mentioned before. Finally, the retrieved candidate answers are encoded into low-dimension distributed representations, denoted as $ AN= \{a_1, a_2, \cdots, a_m\}$, where $m$ is the total number of the words. Moreover, by an interaction between $c^K$ and  $ AN= \{a_1, a_2, \cdots, a_m\}$, we obtain a dense vector $h_a$ as the representation of the retrieved answers,
\begin{equation}
    h_a = W_2 tanh\left(\sum_{i=1}^{m}\left(W_1\left[c^K;a_i\right]\right)\right)
\end{equation}

\subsection{Decoder Module}
We first apply Gated Recurrent Unit~(GRU)~\cite{chung2014empirical}  to obtain the hidden state $h_t$,
\begin{equation}
h_t = \text{GRU} \left(\phi^{emb}(y_{t-1}), h_{t-1}^{*}\right)
\end{equation}
where $\phi^{emb}(\cdot)$ is an embedding function that maps each token to a fixed-dimensional vector.
At the first time step, we use the special symbol ``SOS'' as $y_0$ and the initial hidden state $h_0^{*} = h_a$.
$h_{t-1}^{*}$ consists of three parts, namely, the last hidden state $h_{t-1}$, the attention over $C=\{c^1,c^2, \cdots, c^K\}$ from the encoder module, denoted as $H^c$, and $H^g$, which is calculated by linearly transforming last state $h_{t-1}$ and $h_a$ with a multi-layer perceptron network. We formulate $H^c$ and $H^g$ as follows:

\paragraph{Attention over $\mathbf{C=\{c^1,c^2,\cdots,c^K\}}$} Since MemNN consists of multiple hops, we believe that different hops are relatively independent and have their own semantic meanings over the history dialog. At different time steps, we need to use different semantic information to generate different tokens, so our aim is to get a context-aware representation. We can achieve it by applying attention mechanism to the hidden states achieved at different hops, 
\begin{equation}
    H^c = \sum_{i=1}^K \alpha_{i,t} c^i, \quad \alpha_{i,t} = \frac{e^{\eta(h_{t-1}, c^i)}}{\sum_{i=1}^K e^{\eta(h_{t-1}, c^i)}}
\end{equation}
where $\eta$ is the function that represents the correspondence for attention, usually approximated by a multi-layer neural network.

\paragraph{Template-guided gating mechanism}
As reported in \citet{song2018ensemble}, the top-ranked retrieved reply is not always the one that best match the query, and multiple retrieved replies may provide different reference information to guide the response generation. However, using multiple retrieved replies may increase the probability of introducing ``noisy'' information, which adversely reduces the quality of the response generation. To tackle this issue, we add a gating mechanism 
to the hidden state of candidate answers, aiming at extracting valuable ``information'' at different time steps. Mathematically,
\begin{equation}
    \label{eqn:gate}
    H^g =  \mathit{sigmoid}(h_a \odot h_{t-1}) \odot h_a
\end{equation}
We use element-wise multiplication to model the interaction between candidate answers~($h_a$) and last hidden state of GRU. $h_{t-1}^{*}$ is obtained by concatenating  $h_{t-1}$, $H^c$, and $H^g$.

\paragraph{Hybrid pointer networks} We use another MemNN with three hops for the response generation, where $h_t$ of GRU serves as  the initial reading head, as shown in Figure~\ref{fig:achitecture}. The output of MemNN is denoted as $O=\{o^1,o^2, o^3\}$ and attention weights are $P_o = \{p_o^1, p_o^2, p_o^3\}$.  

Other than a candidate softmax $P_v$ used for generating a word from the vocabulary, we  adopt the idea of Pointer Softmax in~\citet{gulcehre2016pointing},  and introduce an Entity Pointer Networks  (EPN)  and  a  Pattern  Pointer  Networks  (PPN),  where  EPN  is trained to learn  to  copy entity words from dialogue history (or KB), and PPN  is  responsible  for  extracting  pattern words  from  retrieved  answers. For EPN, we use a location softmax $P_h$, which is a pointer network where each of the output dimension corresponds
to the location of a word in the context sequence. Likewise, we introduce a location softmax $P_r$ for PPN. $P_v$ is generated by concatenating the first hop attention read out and the current query vector,
\begin{equation}
    P_v = \textit{softmax}(W_v[o^1;h_t])
\end{equation}

For $P_r$ and $P_h$, we take the attention weights at the second MemNN hop and the third hop of the decoder, respectively: $P_r = p_o^2$ and $P_h = p_o^3$.  The output dimensions of $P_h$ and $P_v$ vary according to the length of the corresponding target sequence.

With the three distributions, the key issue is how to decide which distribution should be chosen to generate a word $w_i$ for the current time step. Intuitively, entity words are relatively important, so we set the selection priority order as $P_r > P_h > P_v$. Instead of using a gate function for selection~\cite{gulcehre2016pointing}, we adopt the sentinel mechanism proposed in~\citet{madotto2018mem2seq}. If the expected word is not appearing in the memories,
then $P_h$ and $P_r$ are trained to produce a sentinel token\footnote{We add a special symbol to the end of each sentence. For example, ``good morning'' is converted to ``good morning \$\$\$''. Therefore, if the model predicts the location of ``\$\$\$'', it means that the expected word is not appearing in the context sequence.}. When both $P_h$ and $P_r$ choose the sentinel token or the masked position, our model will generate the token from $P_v$. Otherwise, it takes the memory content using $P_v$ or $P_r$.

\section{Experimental Settings}
\subsection{Datasets}
We use four public multi-turn task-oriented dialog datasets to evaluate our model, including bAbI~\cite{weston2015towards}, In-Car Assistant~\cite{eric2017copy} , DSTC2~\cite{henderson2014second} and CamRest~\cite{camrest676}. bAbI is automatically generated and the other three datasets are collected from real human dialogs. 

\paragraph{bAbI} 
We use tasks 1-5 from bAbI dialog corpus for restaurant reservation to verify the effectiveness of our model. For each task, there are 1000 dialogs for training, 1000 for development, and 1000 for testing. Tasks 1-2 verify dialog management to check if the model can track the dialog state implicitly.  Tasks 3-4 verify if the model can leverage the KB tuples for the task-oriented dialog system. Tasks 5 combines  Tasks 1-4 to produce full dialogs. 

\paragraph{In-Car Assistant} 
This dataset consists of 3,031 multi-turn dialogs in three distinct domains: calendar sheduling, weather information retrieval, and point-of-interest navigation. This dataset has an average of 2.6 conversation turns and the KB information is complicated. Following the data processing in \citet{madotto2018mem2seq}, we obtain 2,425/302/304 dialogs for training/validation/testing respectively.

\paragraph{DSTC2} 
The dialogs were extracted from the Dialogue State Tracking Challenge 2 for restaurant reservation. Following \citet{bordes2017learning}, we  use merely the raw text of the dialogs and ignore the dialog state labels.  In total, there are 1618 dialogs for training, 500 dialogs for validation, and 1117 dialogs for testing. Each dialog is composed of user and system utterances, and API calls to the domain-specific KB for the user's queries. 

\paragraph{CamRest}
This dataset consists of 676 human-to-human dialogs in the restaurant reservation domain. This dataset has much more conversation turns with 5.1 turns on average. Following the data processing  in \citet{DBLP:conf/eacl/Rojas-BarahonaG17}, we divide the dataset into training/validation/testing sets with 406/135/135 dialogs respectively.

\subsection{Implementation Detail} 
We use the 300-dimensional word2vec vectors to initialize the word embeddings. The size of the GRU hidden units is set to 256. The recurrent weight parameters are initialized as orthogonal matrices. We initialize the other weight parameters with the normal distribution $N(0, 0.01)$ and set the bias terms as zero. We train our model with Adam optimizer~\cite{kingma2015j} with an initial learning rate of $1e-4$. By tuning the hyper-parameters with the grid search over the validation sets, we find the other best settings in our model as follows. The number of hops for the memory network is set to 3, and gradients are clipped with a threshold of 10 to avoid explosion. In addition, we apply the dropout~\cite{hinton2012improving} as a regularizer to the input and output of GRU, where the dropout rate is set to be 0.4. 


\subsection{Baseline Models}

We compare our model with several existing end-to-end task-oriented dialogue systems\footnote{Part of experimental results of baseline models are directly extracted from corresponding published papers.}:
\begin{itemize} 
    \item \textbf{Retrieval method}: This approach directly uses the retrieved result as the answer of the given utterance. Specifically, we use BERT-Base as a feature extractor for the sentences, and we use the cosine distance of the features as our retrieve scores, and then select the one with the highest score.
    \item \textbf{Attn}: Vanilla sequence-to-sequence model with attention~\cite{luong2015effective}.
    \item \textbf{MemNN}: An extended Seq2Seq model where the recurrence read from a external memory multiple times before outputting the target word \cite{sukhbaatar2015end}.
    \item \textbf{PtrUnk}: An augmented sequence-to-sequence model with attention based copy mechanism to copy unknown words during generation \cite{gulcehre2016pointing}.
    \item \textbf{CASeq2Seq}:  This is a copy-augmented Seq2Seq model that learns attention weights to dialogue history with copy mechanism \cite{eric2017copy}.
    \item \textbf{Mem2Seq}: A memory network based approach with multi-hop attention for attending over dialogue history and KB tuples \cite{madotto2018mem2seq}.
    \item \textbf{BossNet}: A bag-of-sequences memory architecture is proposed for disentangling language model from KB incorporation in task-oriented dialogues \cite{raghu2019disentangling}.
    
    \item \textbf{WMM2Seq}: This method adopts a working memory to interact with two separated memory networks for dialogue history and KB entities \cite{chen2019working}.
    
    \item \textbf{GLMP}: This is an augmented memory based model with a global memory pointer and a local memory pointer to strengthen the model's copy ability \cite{DBLP:conf/iclr/WuSX19}.

\end{itemize}

\subsection{Automatic Evaluation Metrics} 

In bAbI dataset, we adopt a common metric per-response accuracy ~\cite{bordes2017learning} to evaluate the model performance. Following previous works~\cite{madotto2018mem2seq}, for three real human dialog datasets, we employ bilingual evaluation understudy (BLEU)~\cite{papineni2002bleu} and Entity F1 scores to evaluate the model’s ability to generate relevant entities from knowledge base and to capture the semantics of the user-initiated dialogue flow~\cite{eric2017copy}. 

\paragraph{BLEU} We use BLEU to measure the n-gram (i.e., 4-gram) matching between the generated responses and the reference responses. The higher BLEU score indicates a better performance of the conversation system.
Formally, we compute the 4-gram precision for the generated response $Y$ as: 
\begin{equation}
P(Y, \hat{Y}) = \frac{\sum_{\tilde{Y}} \text{min}(\eta(\tilde{Y}, Y), \eta(\tilde{Y}, \hat{Y}))}{\sum_{\tilde{Y}} \eta(\tilde{Y}, Y)}
\end{equation}
where $\tilde{Y}$ traverses all candidate 4-grams, $Y$ and $\hat{Y}$ are the ground-truth and predicted responses, $\eta(\tilde{Y}, Y)$ indicates the number of 4-grams in $Y$. After achieving the precision, the BLEU score is then calculated as: 
\begin{equation}
BLEU = \nu(Y, \hat{Y}) \exp (\sum_{n=1}^{4} \beta_n \log P(Y, \hat{Y}))
\end{equation}
where $\beta_n = 1/4$ is a weight score. $\nu(Y, \hat{Y})$ is a brevity penalty that penalizes short sentences. The higher BLEU score indicates better performance of the conversation system. 

\paragraph{Per-response Accuracy}
We adopt the per-response accuracy metric to evaluate the dialog system's capability of generating an exact, correct responses. A generated response is considered right only if each word of the system output matches the corresponding word in the gold response. The final per-response accuracy score is calculated as the percentage of responses that are exactly the same as the corresponding gold dialogues. Per-response accuracy is a strict evaluation measure, which may only be suitable for the simulated dialog datasets. 

\paragraph{Entity F1}
Entity F1 metric is used measure the system's capability of generating relevant entities from the provided task-oriented knowledge base. 
Each utterance in the test set has a set of gold entities. An entity F1 is computed by micro-averaging over all the generated responses.

\begin{table}
    \centering
    \small
    \renewcommand{\arraystretch}{1.3}
    \setlength{\tabcolsep}{1.6mm}{
        \begin{tabular}{c|c|c|c|c|c } 
                \hline
                 Method & BLEU & Ent.F1 & Sch.F1 & Wea.F1 & Nav.F1\\
                \hline
                $R_h^+$ \&  $R_r^+$   & \textbf{12.8} & \textbf{37.8}  & \textbf{50.0}  & \textbf{37.9}  & \textbf{27.5}\\
                $R_h^+$ \& $R_r^-$ & 12.5 &36.1 & 49 & 34.6 & 26.7 \\    
                $R_h^-$ \& $R_r^+$ & 12.3 &36.8  &49.8  &36.6  & 26.1  \\ 
                $R_h^-$ \& $R_r^-$ & 11.6 & 34.8 & 48.3  & 31.8 & 26.5  \\ 
               \hline 
           \end{tabular}
        }
       \caption{Masking comparison experiment on In-Car Assistant. $+$ means with masking and $-$ denotes without. $R_h^+$ \&  $R_r^+$  means that we simultaneously mask NEW and EW in the history dialogue and retrieved answers. }
      \label{tab:mask}
\end{table}

\section{Experimental Results}
\subsection{Automatic Evaluation on Four Datasets}

\begin{table}
    \centering
     \begin{tabular}{c|c|c}
            \hline
            $\theta$ & \# of RA & BLEU \\
            \hline 
            0.3 & 2.48 & 56.1 \\ 
            0.4 & 2.16 & 56.2 \\ 
            0.5 & 1.90 & \textbf{59.8} \\ 
            0.6 & 1.75 & 56.6 \\ 
            1.0 & 1.00 & 56.5 \\
            \hline 
          \end{tabular}
            \caption{Experimental results in terms of BLEU on DSTC2 by using different $\theta$. \# of RA denotes the average number of retrieved answers. \label{tab:LUCENE}}
\end{table}

\paragraph{bAbI} The dataset is automatically generated based on some rules, thus many requests and their corresponding replies are quite similar in terms of the syntactic structure and the wording usage. According to the results shown in Table~\ref{tab:babi}, we can see that our model achieves the best per-response scores in all the five tasks. It is also believed that the retrieved results can contribute to guiding the response generation in this case, which can be inferred from the high threshold value ($\theta = 0.8$).

\begin{table}
    \centering
    \begin{tabular}{c|c|c|c} 
            \hline
            Dataset& BM25 & word2vec & BERT \\
            \hline
            Task1 & 68.7 & 63.1 & 74.8 \\ 
            Task2 & 80.6 & 83.2 & 93.7 \\ 
            Task3 & 83.4 & 77.3 & 80.3 \\
            Task4 & 87.5 & 87.5 & 87.5 \\ 
            Task5 & 82.9 & 66.6 & 83.8 \\ 
            DSTC2 & 45.3 & 37.3 & 47.1\\ 
            CAMREST & 27.7 & 29.0 & 30.9 \\ 
            KVR & 33.5 & 33.7 & 35.3\\ 
          \hline
         \end{tabular}
    \caption{Comparison of different matching methods.}
    \label{tab:retrieval}
\end{table}

\begin{table*}[ht]
\centering

 \begin{tabular}{c|c|c|c|c|c|c|c|c|c}
        \hline
   Task & Retrieval & Attn  &  MemNN & PtrUnk& Mem2Seq  & BossNet & GLMP & WMM2Seq & THPN \\
   \hline

     Task1& 74.8  &100& 99.9 & 100 & 100 &100&100&100 & \textbf{100} \\
 
     Task2& 93.7  &100 & 100 & 100 &100 & 100&100&100& \textbf{100}\\
    
     Task3& 80.3 &74.8 & 74.9 & 85.1 & 94.5 & 95.2 & \textbf{96.3} & 94.9 & 95.8\\

    Task4& 87.5  &57.2 & 59.5 & 100 &100 & 100 & 100 & 100 & \textbf{100}\\
  
     Task5& 83.8 &98.4  & 96.1 & 99.4& 98.2& 97.3 & 99.2 & 97.9 & \textbf{99.6}\\
      \hline  
    \end{tabular}
    
    \caption{Per-response scores on the five tasks of the bAbI dataset with $\theta=0.8$. \label{tab:babi}}
\end{table*}

\paragraph{In-Car Assistant Dataset}  As shown in Table \ref{tab:kvr}, our model achieves all best metrics~(BLEU, Ent.F1, Sch.F1, Wea.F1 and Nav.F1) over other reported models. The possible reason is that the retrieved answers with high relevance to the gold answers provide valid sentence pattern information. By using this sentence pattern information, our model can better control the generation of responses. Additionally, our model improves the success rate of generation correct entities which appeared in the dialogue history.

\begin{table}
\small
\centering
\renewcommand{\arraystretch}{1.2}
 \setlength{\tabcolsep}{1.2mm}{
 
  \begin{tabular}{c|c|c|c|c|c} 
    \hline
      
    Method & BLEU & Ent.F1 & Sch.F1 & Wea.F1 & Nav.F1 \\
    \hline
    Retrieval & 15.3 & 20.1 & 24.9 & 26.3 & 9.4\\
    Attn & 9.3 & 19.9 & 23.4  & 25.6 & 10.8  \\ 
    CASeq2Seq &8.7&13.3&13.4&15.6&11.0\\
     MemNN & 8.3  & 22.7 & 26.9 & 26.7 & 14.9  \\ 
    PtrUnk  &8.3 &22.7 &26.9 &26.7 &14.9 \\
    Mem2Seq &12.6&33.4&49.3&32.8&20.0 \\ 
    BossNet & 8.3 & 35.9 & 50.2 & 34.5 & 21.6 \\
   \hline
   THPN &\textbf{12.8}& \textbf{37.8}  & \textbf{50.0}  &\textbf {37.9} & \textbf{27.5}\\ 
    \hline
  \end{tabular}
  }
  \caption{Evaluation results on the In-Car Assistant dataset with $\theta=0.3$.\label{tab:kvr}}
\end{table}

\begin{table*}
\centering \small 
  \renewcommand{\arraystretch}{1.3}
  \setlength{\tabcolsep}{1.6mm}{
     \begin{tabular}{c|c|c|c|c|c|c|c|c}
            \hline
       \multirow{2}{*}{Task} &Task1&Task2&Task3&Task4&Task5&DSTC2&DSTC2&DSTC2\\
       \cline{2-9}
       & (BLEU) &(BLEU) & (BLEU) &  (BLEU) &  (BLEU) &  (BLEU)& (F1) & (Per-Res)\\
       \hline
         THPN &100 & 100&98.9& 100 & 99.9 & 59.8 & 76.8 
         &47.7\\
         W/O IR&100 &100 &96.5 & 100 & 99.2 & 57.8 &73.2  &45.9 \\
         W/O Ptr&100 & 100&97.7& 89.9 & 98.5 & 58.1 &72.6 & 46.1 \\
         W/O Gate& 100 & 100&95.9& 94.4 & 99.2 & 57.7 & 74.1  &45.8\\
          \hline  
        \end{tabular}
    }
    \caption{Ablation test results of our THPN model on bAbI and DSTC2 datasets. \label{tab:ablation}}
\end{table*} 

\paragraph{DSTC2 and CamRest Datasets} We also present the evaluation on DSTC2 and CamRest datasets in Table~\ref{tab:dstc} and Table~\ref{tab:camrest}, respectively. By comparing the results, we can notice that our model performs better than the compared methods. On the DSTC2, our model achieves  the state-of-the-art performance in terms of both Entity F1 score and BLEU metrics, and has a comparable per-response accuracy with  compared methods. On the CamRest, our model obtains the best Entity F1 score but has a drop in BLEU in comparison to Mem2Seq model.

\begin{table}
    \centering
    \begin{tabular}{c|c|c} 
            \hline
                Method& Ent.F1 & BLEU \\
                \hline
                Retrieval & 21.1& 47.1\\
                Attn & 67.1 & 56.6  \\ 
                KV Net & 71.6& 55.4  \\ 
                Mem2Seq & 75.3 & 55.3  \\ 
                GLMP & 67.4 & 58.1 \\
                \hline
               THPN  & \textbf{76.8} & \textbf{59.8} \\ 
           \hline
         \end{tabular}
         \caption{Evaluation on DSTC2($\theta=0.5$).}
         \label{tab:dstc}
\end{table}

\begin{table}
    \centering
       \begin{tabular}{c|c|c} 
            \hline
            Method & Ent.F1 & BLEU  \\
            \hline
            Retrieval & 7.9& \textbf{21.2}\\
            Attn & 21.4 & 5.9    \\ 
            PtrUnk & 16.4& 2.1    \\ 
            KV Net &9.1 &4.3\\
            Mem2Seq & 27.7 & 12.6   \\ 
           \hline 
           THPN  & \textbf{30.9} & 12.9   \\ 
            \hline
         \end{tabular}
         \caption{Evaluation on CamRest($\theta=0.4$).}
         \label{tab:camrest}
\end{table}

\subsection{Ablation Study}
An ablation study typically refers to removing some components or parts of the model, and seeing how that affects performance. To measure the influence of the individual components, we evaluate the proposed THPN model with each of them removed separately, and then measure the degradation of the overall performance. 
Table~\ref{tab:ablation} reports ablation study results of THPN on bAbI and DSTC2 datasets by removing retrieved answers (w/o IR), removing EPN and PPN in decoding (w/o Ptr), removing  answer-guided gating mechanism (w/o Gate), respectively. For example, ``w/o Gate'' means we do not use the answer-guided gating mechanism while keeping other components intact.

If the retrieved answer is not used, the performance reduces dramatically, which can be interpreted that without the guiding information from the retrieved answer, the decoder may deteriorate quickly once it produce a ``bad'' word since it solely relies on the input query.

If no copy mechanism is used, we can see that Entity F1 score is the lowest, which indicates that many entities are not generated since these entity words may not be included in the vocabulary. Therefore, the best way to generate some unseen words is to directly copy from the input query, which is consistent with the findings of previous work~\cite{eric2017key,madotto2018mem2seq}.

If the gate is excluded, we can see around $2$\% drop for DSTC2. A possible reason is that some useless retrieved answers introduce ``noise'' to the system, which deteriorates the response generation.

 \subsection{Effect of Masking Operation}\label{sec:mask}



 
 To validate the effectiveness of the masking operation, we carry out a comparison experiment on In-Car Assistant, and present the results in Table~\ref{tab:mask}. From Table~\ref{tab:mask}, we can see that $R_h^+$ \& $R_r^+$ achieves the best performance while $R_h^-$ \& $R_r^-$ has the lowest scores. By diving into the experimental results, we find that if we do not mask EW in the retrieved answers, the model copies many incorrect entities from the retrieved answers, which reduces the Entity F1 scores. If we do not mask NEW in the history dialogue, the percentage of NEW copied from the history dialogue is high, most of which are unrelated to the gold answer, thus bringing down the BLEU score.

\subsection{Analysis on Retrieved Results}

\paragraph{Comparison of Different Retrieval Methods}
According to our preliminary experimental results, we observed that better retrieved candidate answers could further improve the overall model performance in response generation. Therefore, we also conduct experiments to evaluate the effectiveness of three popular text matching methods, including  BM25~\cite{robertson2009probabilistic}, word2vec~\cite{mikolov2013efficient} and BERT~\cite{devlin2018bert}. Here, BLEU is utilized as our evaluation criterion. From the experimental results shown in Table~\ref{tab:retrieval}, we can see that using BERT~\cite{devlin2018bert}, a transformer-based pre-trained language model, achieves the highest BLEU scores. A possible reason is that  the size of each training dataset is limited, the word co-occurrence based algorithms~(e.g., BM25) may not capture the semantic information, thus result in poor retrieving performance. 

\paragraph{One vs. Multiple Retrieved Answers}
Cosine similarity is not an absolute criterion and there is no guarantee that a candidate with higher cosine value will always provide more reference information to the response generation. Therefore, we conduct an experiment to investigate the effect of the number of retrieved answers. By setting different cosine threshold values $\theta$, we retrieve different numbers of answer candidates. In particular, if no answer candidate satisfies the given threshold, we choose one with the highest cosine value. To limit the number of retrieved answers, we only select the top-$3$ results if there are more than three answer candidates that have higher consine values than the given threshold $\theta$. 

Table~\ref{tab:LUCENE} gives the experimental results of DSTC2 dataset under different threshold $\theta$ values. When $\theta$ is set to be $1.0$, it is considered as a special case where only one answer is retrieved.  We  can observe that using multiple answer candidates obtains higher performance than only using one result.  It is intuitive that the model will be misguided if the retrieved single answer has no relation to the given request, and using multiple candidate answers can ameliorate this issue. 



\paragraph{Setting of $\theta$}
Although using more retrieved answers might improve the chance of including the relevant information, it may also bring more ``noise'' and adversely affect the quality of retrieved answers.  From Table \ref{tab:LUCENE}, we can see that with the reduced value of $\theta$, the average number of retrieved candidate answers increase, but the model performance does not improve accordingly. Experimental results on the other datasets demonstrate that the $\theta$ is not fixed and needs to be adjusted according to the experimental data.

\section{Conclusion}
In  task-oriented  dialog  systems,  the  words  and sentence structures are relatively limited and fixed, thus it is intuitive that the retrieved results can provide valuable information in guiding the response generation.  In this paper, we retrieve several potentially relevant answers from a pre-constructed domain-specific conversation repository as guidance answers, and incorporate the guidance answers into both the encoding and decoding processes. We copy the  words from the previous context  and the retrieved answers directly, and generate words from the vocabulary.  Experimental results  over four datasets have demonstrated the effectiveness of our model in generating informative responses.
In the future, we plan to leverage the dialogue context information to retrieve candidate answers turn by turn in multi-turn scenarios. 

\bibliographystyle{acl_natbib}
\bibliography{acl2021}

\end{document}